\documentclass[11pt,a4paper]{article}
\usepackage[margin=2.5cm]{geometry}
\usepackage{amsmath,amsfonts,amssymb}
\usepackage[pdftex]{graphicx}
\usepackage{mathptmx} 

\usepackage[colorlinks=true,linkcolor=blue,citecolor=blue]{hyperref}

\usepackage{booktabs, multirow}
\usepackage{siunitx}
\usepackage{gensymb}
\usepackage{subfigure}
\usepackage{dcolumn}
\usepackage{url}
\usepackage{setspace}
\usepackage{makecell}  

\begin{document}

\def\papertitle{Two-stage deep learning framework for the restoration of incomplete-ring PET images}
\title{\papertitle}

\author{
    Yeqi Fang\thanks{Email: fangyeqi@stu.scu.edu.cn}  and Rong Zhou \\[0.3em]
    \small College of Physics, Sichuan University, Chengdu 610065, China
}

\date{}
\maketitle
\vspace{-1cm}
\begin{abstract}
Positron Emission Tomography (PET) is an important molecular imaging tool widely used in medicine. Traditional PET systems rely on complete detector rings for full angular coverage and reliable data collection. However, incomplete-ring PET scanners have emerged due to hardware failures, cost constraints, or specific clinical needs. Standard reconstruction algorithms often suffer from performance degradation with these systems because of reduced data completeness and geometric inconsistencies.
We present a two-stage deep-learning framework that, without incorporating any time-of-flight (TOF) information, restores high-quality images from data with about 50\% missing coincidences—double the loss levels previously addressed by CNN-based methods. The pipeline operates in two stages: a projection-domain Attention U-Net first predicts the missing sections of the sinogram by leveraging spatial context from neighbouring slices, after which the completed data are reconstructed with OSEM algorithm and passed to an cascaded U-Net \& warm-start diffusion model for image refinement.
This module starts the reverse diffusion process from the U-Net coarse prediction 
rather than pure Gaussian noise.
Using 613 simulated brain volumes from real scans (196 healthy brain samples, 217 Alzheimer’s disease samples, and 200 Mild Cognitive Impairment samples), the result shows that our model successfully preserves most anatomical structures and tracer distribution features with PSNR of 38.18 to 38.59 dB and SSIM of 0.9904 to 0.9925.
Our two-stage deep‐learning framework effectively restores high‐quality PET images from over 50 \% incomplete‐ring data, achieving near‐complete anatomical fidelity and robust performance without requiring TOF information.
\end{abstract}  

\noindent\textbf{Keywords:} Incomplete-ring, Sinogram restoration, Attention U-Net, Diffusion-based refinement


\section{Introduction}
\label{chap:introduction}

Positron emission tomography (PET) is a powerful molecular imaging technique that provides quantitative visualization of metabolic processes within living tissue \cite{townsend2004, muehllehner2006positron, lameka2016positron, shukla2006positron, nutt2002history}. PET operates on the principle that positrons annihilate upon encountering electrons, producing detectable 511keV photon pairs: a radiotracer introduces positrons that annihilate upon encountering electrons, and then produce pairs of $511$keV photons traveling in opposing directions. 
The photon pairs are then detected by a pair of scintillation sensors, and each coincidence event will define a line-of-response (LOR).
Traditional PET systems usually use complete 360$^\circ$ detector rings to maximize their sensitivity and angular coverage \cite{townsend2004}.

However, incomplete-ring systems have emerged from practical necessity rather than theoretical preference: they reduce costs and complexity, allow closer access to patients in specialized applications like breast scanning or in-beam therapy monitoring \cite{surti2008}, create "open" configurations that alleviate claustrophobia \cite{tashima2012, krishnamoorthy2021}, and enable novel applications such as dual-panel brain imaging systems \cite{zhang2020}. The trade-off is substantial—incomplete angular coverage creates gaps in projection data, turning reconstruction into an underdetermined problem. Missing angular views inevitably introduce artifacts and resolution non-uniformity \cite{kak1988, surti2008}. Even time-of-flight capabilities, which partially fix these problems, can't fully compensate for limited view angles \cite{surti2008, krishnamoorthy2021}.

Researchers have investigated various methods to achieve progressive improvements in incomplete-ring PET reconstruction. Analytical ways like filtered back-projection falter immediately—their assumption of complete data leads to pronounced streak artifacts \cite{kak1988}. Iterative approaches like maximum-likelihood expectation-maximization (MLEM) are better by modeling the acquisition process \cite{qi2006}, but they still have artifacts along missing view directions \cite{zhang2020}.
Penalized likelihood methods incorporating prior constraints show promise, as Zhang \textit{et al.} demonstrated by using high-quality prior images to enhance contrast recovery in a dual-panel head-and-neck PET system \cite{zhang2020}. And sinogram completion approaches attempt to estimate or interpolate the absent projection data prior to reconstruction \cite{makkar2024partial}. However, these methods remain insufficient to solve the fundamental problem. 

Deep learning has shown remarkable success in many fields \cite{PhysRevD.110.063011, liu2024deep, reader2020deep}. Regression-based networks have shown promise in low-dose PET denoising and partial data reconstruction \cite{Kandarpa_2021} by learning direct mappings between degraded and high-quality images. Liu \textit{et al.}'s U-Net approach to transform artifact-degraded partial-ring PET images \cite{liu2019} demonstrated initial success, but the model in this work is mainly targeted for mild data loss and training data didn't consider the context of sinograms.
However, current available techniques have some drawbacks.
For examples, while GAN approaches are good at generating visually compelling reconstructions, \cite{xue2023cg3dsrganclassificationguided3d} they are unstable when training and often highly sensitive to hyperparameters, which limit their clinical use.
Methods based on explicit likelihood modeling—like VAEs and flow-based approaches—offer solid theoretical foundations. But they usually lose details in reconstructed images and their low speeds also make them not suitable for clinical use.
Model-based deep learning frameworks and generative models \cite{reader2023, vashistha2024} show promise but risk introducing hallucinated features—a dangerous proposition in medical imaging.

Restoring images from  incomplete projection data is a common challenge across tomographic modalities. In limited-angle CT, deep learning remains effective with angular ranges of 30--90$^\circ$~\cite{lahiri2019lose, germer2023limited}. Networks working in the sinogram domain can also recover gaps more than half the full
coverage~\cite{zhao2019sinogram}. Similar progress in accelerated MRI has shown that deep learning can still recover clinically useful images despite large missing fractions
~\cite{wang2024knowledge}. As for incomplete-ring PET, however, prior CNN-based methods addressed at most $\sim$25\% coincidence loss~\cite{liu2019partial}.
Unsupervised approaches have been validated only on mild geometric
gaps~\cite{shan2024dip, makkar2024inr}. This work extends this
line of research to $\sim$50\% coincidence loss. To our knowledge, this work
provides the first systematic benchmark at this severity on a dataset
of 613 brain volumes.

While incomplete-ring PET systems find their primary applications in specialized scenarios such as breast imaging and in-beam therapy monitoring, brain PET imaging also presents compelling use cases for partial-ring designs. Dual-panel brain imaging systems offer reduced claustrophobia for patients while maintaining diagnostic quality. Additionally, open-configuration brain scanners enable better patient access for interventional procedures and real-time monitoring during treatments. 
More importantly, brain PET serves as an ideal testbed for developing and validating incomplete-ring reconstruction algorithms due to its well-characterized anatomy and the availability of standardized datasets. The complex cerebral structures provide a rigorous evaluation framework for sinogram completion methods, while the high contrast between gray matter, white matter, and CSF regions allows for sensitive assessment of reconstruction artifacts. The techniques developed and validated on brain data can subsequently be adapted to other anatomical regions where incomplete-ring designs offer specific clinical advantages.


In this work we introduce a two-stage deep-learning framework that 
restores highly degraded incomplete-ring PET data. To our knowledge, 
it is the first framework validated under $>$50\% coincidence loss. 
Our contributions are four-fold:
(1) a five-channel slice-stacking Attention U-Net for sinogram 
restoration under severe angular deficits;
(2) a warm-start Resfusion-based diffusion module~\cite{shi2023resfusion} 
that refines OSEM images by learning the residual from the U-Net 
coarse prediction;
(3) a systematic benchmark on 613 simulated brain volumes (healthy, AD, MCI) 
at $\sim$50\% coincidence loss;
and (4) a thorough ablation study and comparison against existing 
incomplete-ring PET reconstruction methods.

The remainder of this thesis is organized as follows:
\textbf{Section~\ref{chap:background}} provides an overview of PET and its basic physics like principles of coincidence detection, and the main problems caused by incomplete rings. We also summarize classical and modern reconstruction methods.
\textbf{Section~\ref{chap:methods}} presents our contribution in full: the simulation framework for complete and partial rings, the projection-completion Attention U-Net, the subsequent OSEM reconstruction, and the diffusion-based image-refinement module, together with implementation and training details.
\textbf{Section~\ref{chap:results}} reports quantitative and qualitative evaluations on 613 brain volumes under several angular-loss patterns, comparing sinogram restoration, preliminary OSEM images and final refined outputs.
\textbf{Section~\ref{chap:conclusion}} summarizes our findings by analysing metrics of model performance on test dataset, and discusses limitations and directions for future improvements.
\textbf{Appendices} provide additional experiments, extended qualitative results, and tables of hyperparameters used in the study.

\section{System Model and Dataset}

\label{chap:background}

\begin{figure*}[htbp]
    \centering
    \vspace{-0.2cm}
    \includegraphics[width=0.34\textwidth]{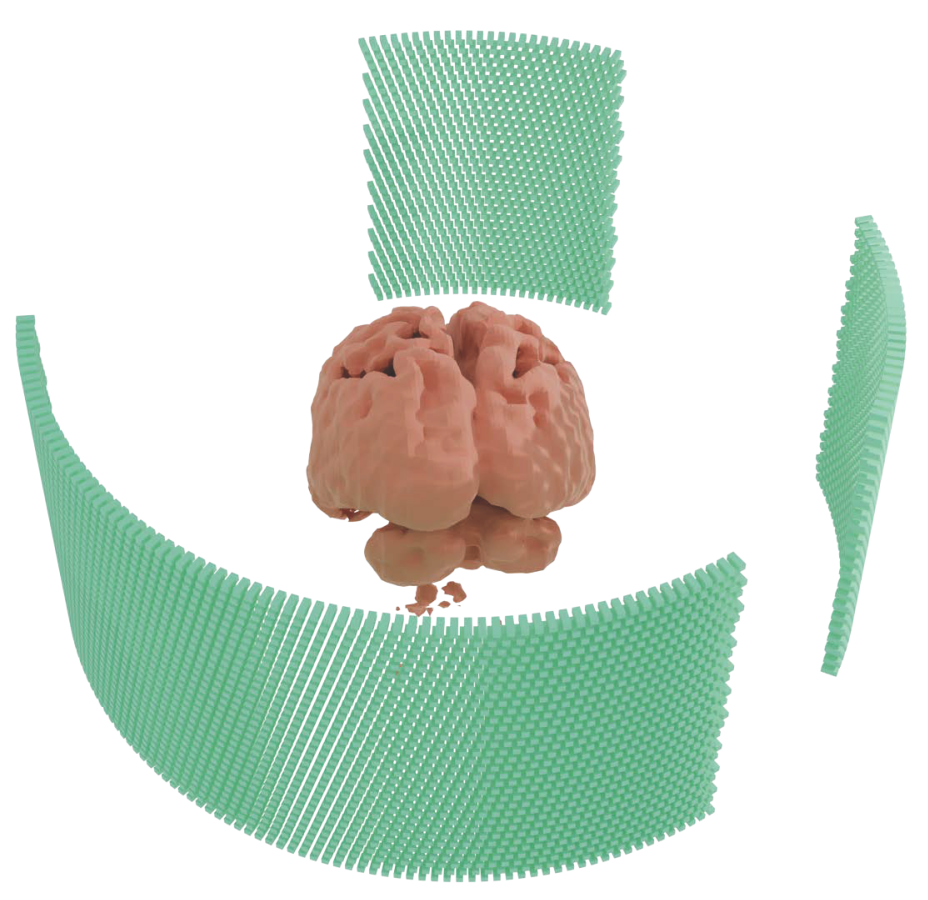}
    \includegraphics[width=0.34\textwidth]{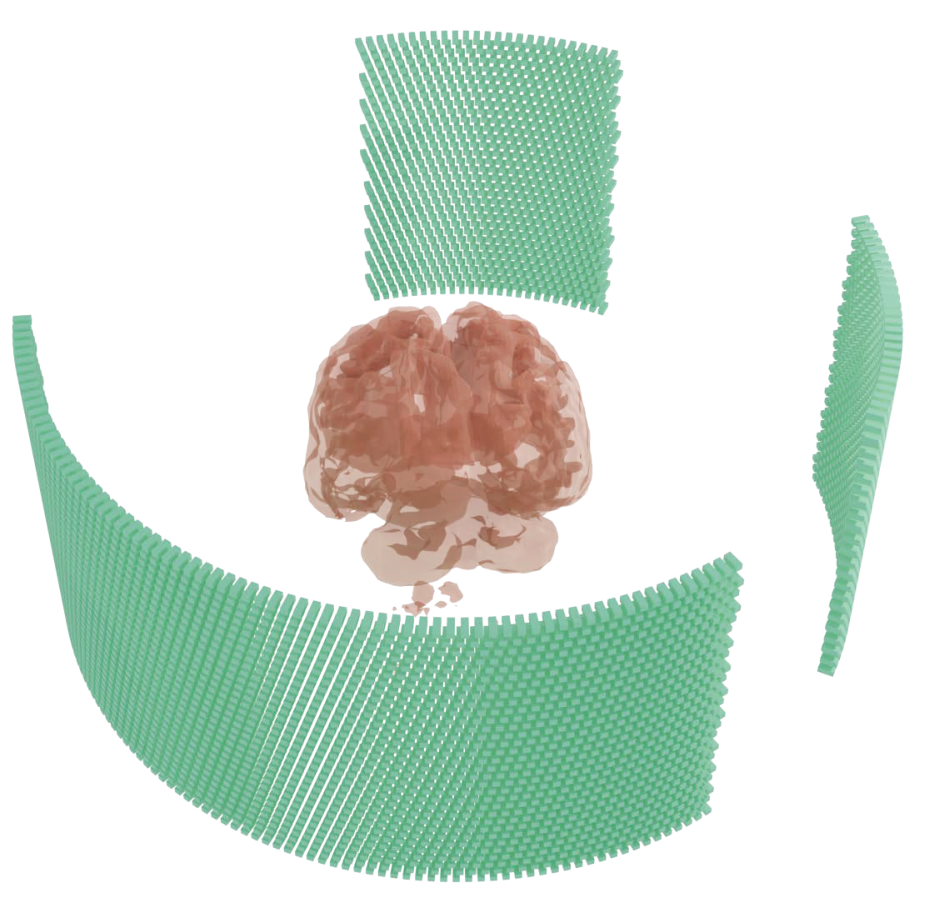}
    \includegraphics[width=0.34\textwidth]{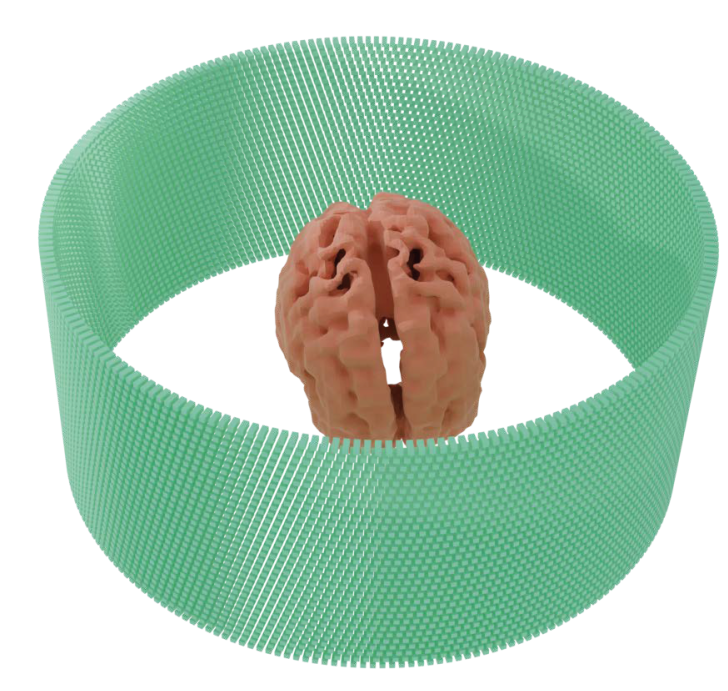}
    \includegraphics[width=0.34\textwidth]{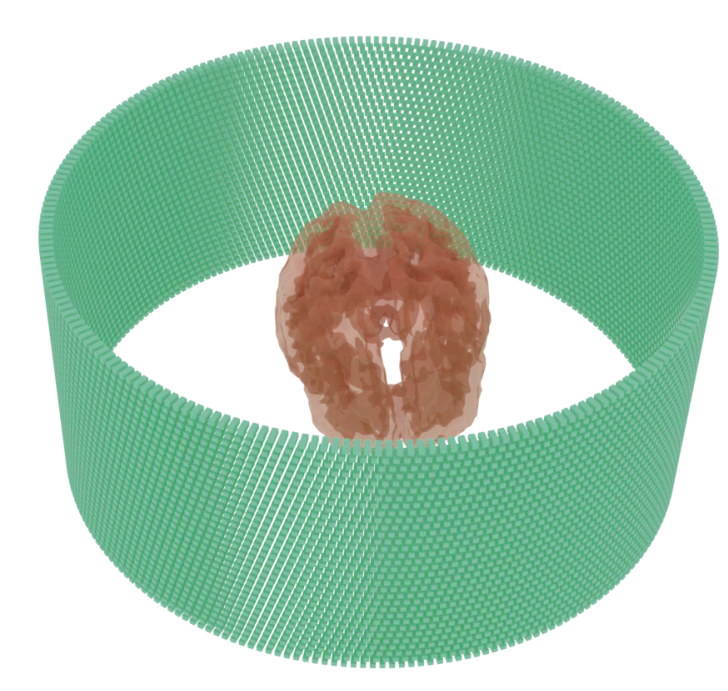}
    \vspace{-0.2cm}
    \caption{Three-dimensional schematic of incomplete and complete ring PET scanner detector structure, with the right image showing a perspective view.}
    \vspace{-0.2cm}
    \label{fig:pet_structures2}
\end{figure*}

Incomplete ring configurations arise from practical constraints including cost limitations, hardware failures, and specific clinical requirements. Budget constraints often drive this compromise—each detector module constitutes a substantial cost, and reducing their number can make PET technology accessible to more facilities. In specialized applications, partial ring designs actually provide benefits: breast imaging benefits from closer detector positioning, interventional procedures require open scanner designs, and claustrophobic patients experience less anxiety with more open configurations.

To deal with these problems systematically, we made a thorough simulation framework modeling both complete and incomplete PET geometries. Our approach begins with a standard ring configuration of $R$ axial rings (we used $R=42$ in our experiments), each containing $D=182$ detectors arranged cylindrically. The scanner radius ($\rho$) of 253.71 mm balances spatial resolution against sensitivity.

From listmode to sinogram, and from sinogram to reconstruction, the information is constantly losing. So, rather than attempting direct image-domain recovery, we pivoted to a two-stage method—first converting listmode data to sinograms with missing sections, then applying specialized techniques to complete these sinograms before final reconstruction. This approach, while computationally more intensive, provided a more applicable way to solve the missing data problem by retaining as much information as possible from sinogram.
For simulation purposes, we mapped a $80\times128\times128$ voxel grid to physical space, yielding in-plane voxel dimensions of approximately 2.78 mm.

Table \ref{tab:detector_params} summarises the target configuration of the in‑house PET prototype, an incomplete ring system that is currently being assembled in our laboratory.  Because several mechanical layouts are still under evaluation, we designed a custom geometry interface for the Monte‑Carlo toolkit SimSET\cite{Harrison157P}.  The interface allows any combination of inactive rings or angular gaps to be enabled or disabled at run‑time, giving us maximal flexibility while the hardware design is finalised.
Once the full‑ring list‑mode data have been generated, we post‑process the coincidences to emulate incomplete geometries: selected detector rings or azimuthal sectors are masked out, and any coincidence that involves at least one inactive crystal is discarded.  This two‑stage strategy keeps the underlying activity distribution, Poisson statistics and scatter model identical between the reference and degraded datasets, isolating the effect of missing angular information.
Importantly, detector-specific sensitivity variations were incorporated by applying crystal-wise efficiency factors (Gaussian spread $\approx 10\%$ around the nominal sensitivity), ensuring that the coincidence statistics also reflect non-uniform detector response. 
The attenuation effect of seven tissue types (i.e, brain, bone, muscle, fat and blood) was considered in the simulations and then corrected during the image reconstruction.

\begin{table}[htbp]
    \centering
    \caption{PET Scanner Configuration Parameters (based on our planned experimental prototype for future validation studies)}
    \label{tab:detector_params}
    \begin{tabular}{l l}
    \hline \hline \addlinespace[2pt]
    \textbf{Parameter} & \textbf{Value} \\
    \hline\addlinespace[2pt]
    Radius & 253.71 mm \\
    Crystal transaxial spacing & 4.02 mm \\
    Crystal axial spacing & 5.37 mm \\
    Module axial spacing & 37.56 mm \\
    Crystal elements (transaxial) & 13 \\
    Crystal elements (axial) & 7 \\
    Transaxial sectors & 28 \\
    Axial sectors & 1 \\
    Modules (axial) & 6 \\
    Crystals per ring & 182 \\
    Number of rings & 42 \\
    \hline \hline
    \end{tabular}
\end{table}

We used a dataset of 613 brain PET simulated  volumes whose activity distributions were obtained from the 
Alzheimer's Disease Neuroimaging Initiative (ADNI) 
database~\cite{weiner2017adni}, comprising 196 healthy controls, 
217 Alzheimer's disease (AD) subjects, and 200 Mild Cognitive 
Impairment (MCI) subjects. Each volume is resized to $80\times128\times128$ (axial $\times$ coronal $\times$ sagittal) 
with voxel size of $2.78\times2.78\times2.78$\,mm$^3$.

. The study was retrospective. The use of patient information will not adversely affect them, so we waived informed consent, but data confidentiality was ensured. All methods strictly adhered to relevant guidelines and regulations.
Each image simulates 2 billion events including both true and scatter events, the detectors detect 560 million events on average and record them in the listmode data.

The complete ring configuration (Figure~\ref{fig:pet_structures2}) provides our reference reconstruction $\mathbf{Y}_A$, representing the best-case scenario with full angular sampling. For incomplete ring experiments, we selectively remove detectors—either entire rings or angular segments—and generate a degraded reconstruction $\mathbf{Y}_B$ from the resulting partial data. The striking quality difference between $\mathbf{Y}_A$ and $\mathbf{Y}_B$ shows the reconstruction challenge we aim to solve.

\begin{figure*}[ht]
    \centering
    \includegraphics[width=\textwidth]{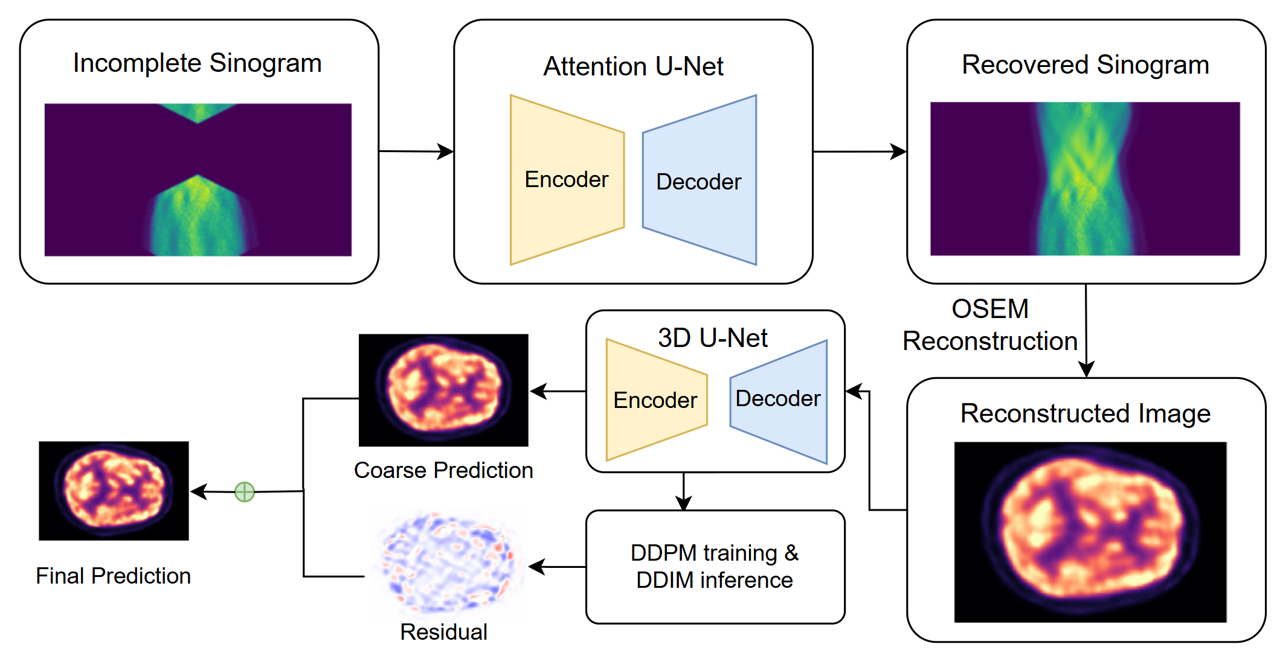}
    \vspace{-.5cm}
    \caption{Overall flow diagram of the incomplete ring PET reconstruction method. First, incomplete sinograms due to missing detectors are repaired through a U-Net model to generate complete sinogram data . Then the complete sinogram is converted into a PET image using the OSEM iterative reconstruction algorithm. Finally, the image is refined using a U-Net \& diffusion model to remove residual artifacts and enhance high-frequency details. 
  }
    \vspace{-.2cm}
    \label{fig:reconstruction_workflow}
\end{figure*}

\section{Methods}
\label{chap:methods}

As shown in Figure \ref{fig:reconstruction_workflow}, this study proposes an innovative incomplete ring PET reconstruction framework that effectively addresses the data incompleteness problem caused by missing detectors through a multi-stage strategy. In the first stage, the system first processes incomplete sinograms (top left) resulting from missing detectors, completing the missing data through a trained optimized U-Net deep learning model to generate complete sinograms (top right). This sinogram restoration process fully utilizes the five-channel input strategy, effectively integrating spatial and temporal context information.

\subsection{Sinogram Reconstruction Based on Attention U-Net}
\label{sec:unet_sinogram}

\begin{figure*}[htbp]
\centering
\vspace{-.3cm}
\includegraphics[width=0.96\textwidth]{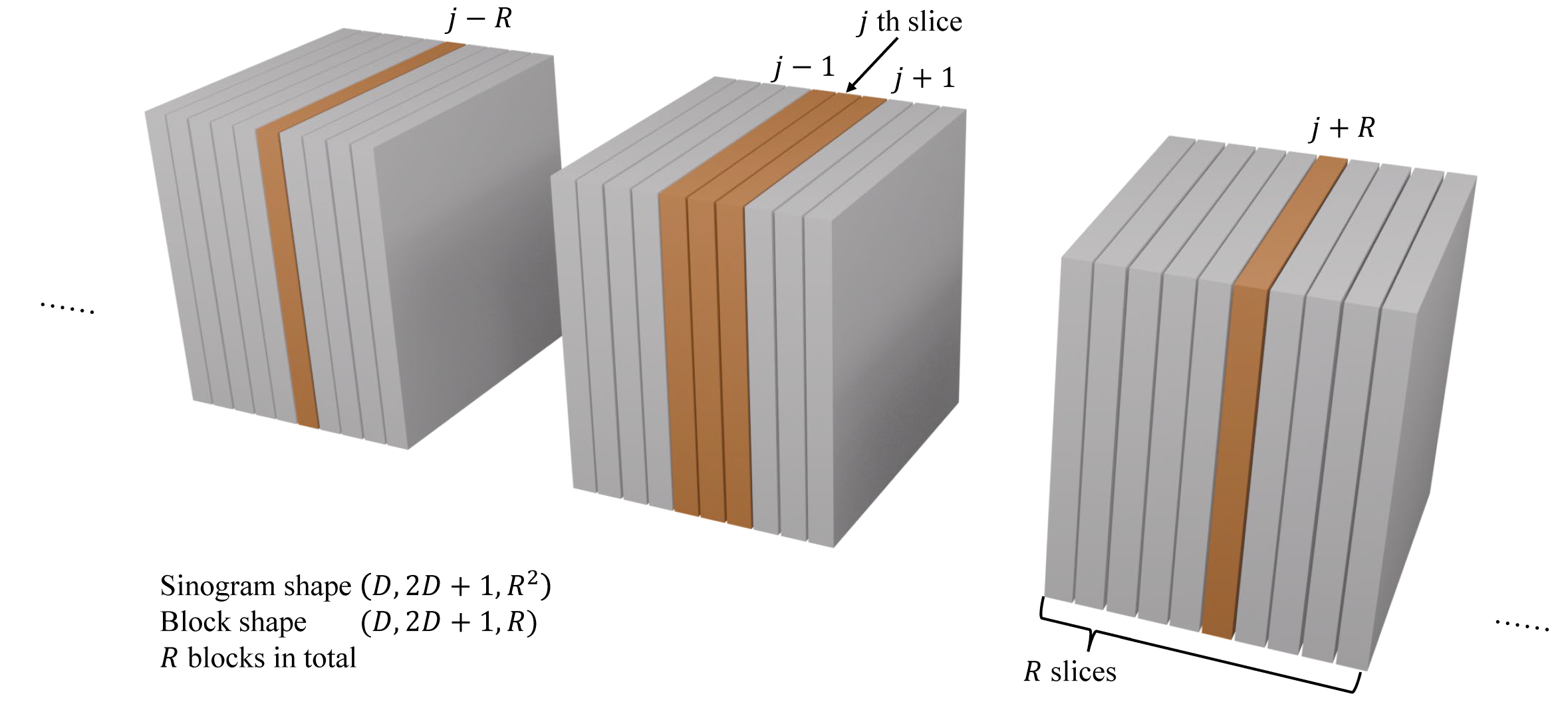}
\vspace{-.3cm}
\caption{Visualization of the five-channel tensor preparation of sinogram data for training. Each cube represents a sinogram slice block with dimensions $(D, 2D+1, R^2)$. The orange highlighted parts are the selected slices ($j-R, j-1, j, j+1, j+R$), which form the five-channel input for the restoration model, showing the spatial and temporal relationships captured in each input tensor.}
\label{fig:sinogram_structure}
\end{figure*}

Sinogram repair is fundamental in incomplete ring PET imaging, as its quality directly affects the precision of reconstructed images. 
The shape of sinogram tensors is $(D, 2D+1, R^2)$, or $(182, 365, 1764)$ in our work, where $R$ is the number of axial rings and $D$ is the number of detectors each ring. It is evident that this tensor is too large to be directly fed into any UNet-like models. Therefore, we divided this large $(D, 2D+1, R^2)$ tensor into $R\times(D, 2D+1, R)$ tensors. 
As shown in Figure~\ref{fig:sinogram_structure}, the central channel of each input tensor corresponds to the current sinogram slice, while the direct spatial neighbors (slices $j-1$ and $j+1$) and temporal neighbors from previous and subsequent sinogram periods (slices $j-R$ and $j+R$) constitute the other four channels. For boundary handling, when adjacent indices exceed the dataset range, the central slice itself is used for channel filling, ensuring input dimension consistency. These neighboring slices are important for improving the model's understanding of local structural continuity and temporal consistency. Consequently, we obtain $1764\times(182, 365, 5)$ tensors for each sinogram. 
To make sure each image was evaluated once in the test set, we implemented 6-fold cross-validation across the 613 volumes. The predictions obtained from our model were reconstructed using the OSEM algorithm to achieve preliminary geometric recovery, and these reconstructions were subsequently used as inputs for our second stage framework.

Although traditional U-Net performs excellently in medical image segmentation and reconstruction tasks \cite{ronneberger2015unetconvolutionalnetworksbiomedical}, it lacks the ability to selectively focus on key features when dealing with complex incomplete ring PET sinogram restoration problems. The Attention U-Net \cite{oktay2018attentionunetlearninglook} adopted in this study enhances the model's perception of important feature regions through spatial attention mechanisms while suppressing the influence of irrelevant features, which is more useful for restoring sinograms from incomplete data. 

We also tested some transformer-based U-Net like UNETR \cite{hatamizadeh2021unetrtransformers3dmedical}, which combines ViT encoder and convolutional decoder, and TransUNet \cite{chen2021transunettransformersmakestrong}, where both encoder and decoder are completely based on transformers. They are trained and tested under the same training protocol for comparison.

The main innovation of Attention U-Net is the introduction of attention gating (AG) modules in the skip connection path of the original U-Net. These AG modules can adaptively highlight significant structures in the feed-forward feature maps while suppressing less relevant regions. 
For sinogram reconstruction tasks, this mechanism is especially useful as it can selectively focus on structural features around missing areas, and then more accurately infer missing angular data. The mathematical expression of attention gating can be described as \cite{oktay2018attentionunetlearninglook}:
\begin{equation}
\alpha_i^l = \sigma_2(\psi^T(\sigma_1(W_x^T x_i^l + W_g^T g_i + b_g)) + b_\psi),
\end{equation}
where $x_i^l$ is the low-level feature from the encoder, $g_i$ is the gating signal from the decoder (high-level feature), $\sigma_1$ and $\sigma_2$ are ReLU and Sigmoid activation functions respectively, and $W_x$, $W_g$, $b_g$, and $b_\psi$ are learnable parameters. $\alpha_i^l \in [0,1]$ is the calculated attention coefficient used to control the importance of features.
After processing through the attention gate, the features can be represented as:
\begin{equation}
\hat{x}_i^l = x_i^l \cdot \alpha_i^l.
\end{equation}

The Attention gates learn to focus on relevant regions of the encoder feature maps by assigning weights based on the context, particularly in cases of severe angular loss. AGs can create a stronger understanding of sinogram continuity and consistency.
After testing on our dataset, the results showed that Attention U-Net performs better with an average increase of 1.24dB in PSNR and 0.052 in SSIM compared to original U-Net in sinogram reconstruction tasks. This means that attention mechanism is more effective in processing incomplete ring PET data.

Model training uses the Adam optimizer with an initial learning rate of $10^{-4}$ and adds   $10^{-5}$ weight decay to prevent overfitting. 
Training efficiency is enhanced through mixed precision computation and gradient scaling techniques, combined with a ReduceLROnPlateau dynamic learning rate scheduler (with a decay factor of 0.3 when validation loss shows no improvement for 3 consecutive epochs), achieving stable convergence.

\subsection{Image Refinement Based on 3D U-Net and Diffusion Model}
\label{sec:diffusion_model}
Then, to further improve the image quality in geometry space, we have the second stage of our framework, which is a cascaded dual-model architecture. The 3D Attention U-Net model takes the OSEM reconstruction as input and produces a coarse prediction IP. 
The diffusion model then takes IP as the warm-start condition. Rather 
than starting from pure Gaussian noise, the forward process is defined as~\cite{shi2023resfusion}
\begin{equation}
    \mathbf{x}_t = \mathrm{IP} + \sqrt{\bar\alpha_t}
    (\mathbf{x}_0 - \mathrm{IP}) +
    \sqrt{1-\bar\alpha_t}\,\boldsymbol{\epsilon},
\end{equation}
so the reverse process starts from a noisy version of IP.
The denoiser 
inside the diffusion model is a separate U-Net backbone, 
which takes the concatenation of IP and the noisy image $\mathbf{x}_t$ 
as input at each timestep and learns to predict the residual 
$\mathbf{x}_0 - \mathrm{IP}$.
The final output is therefore 
IP plus the predicted residual. The two U-Nets are strictly 
separate components. The model is trained from scratch using an 
MSE loss with min-SNR-$\gamma$ weighting~\cite{hang2023minsnr}.

The simulation framework and model training were implemented in Python 3.10 
using PyTorch 2.6.0+cu124. The Deep learning source code is publicly available at 
\url{https://github.com/Yeqi-Fang/Incomplete-Ring}.
The simulation framework is publicly available at 
\url{https://github.com/Yeqi-Fang/PET-Simulator}. 
Inference on a single $80\times128\times128$ volume takes 
approximately 18.6\,s on a single NVIDIA RTX 4090 GPU and 16 vCPU Intel(R) Xeon(R) Platinum 8352V CPU @ 2.10GHz, with
sinogram restoration 1.7\,s, OSEM reconstruction 13.5\,s, 
and diffusion refinement 3.4\,s for about 20 DDIM steps.

\section{Results}
\label{chap:results}
In order to investigate the effect of the form of loss on the reconstruction results while ensuring that the total amount of data loss remained constant, we varied the range of angles at which the data were missing and ensured that the total range of missing angles remained constant and repeated the experimental procedure. The results on test set are listed in Table~\ref{tab:results}. 
The table includes PSNR and SSIM values of the recovered sinogram data by the first stage Attention U-Net and the final recovered images using the two-stage model. 


\begin{table}[htbp]
    \centering
    \caption{Quantitative results for different angular-loss patterns}
    \label{tab:results}
    \begin{tabular}{llll}
    \hline \hline \addlinespace[2pt]
    \textbf{Lost angle range}&\textbf{Parameter} & \textbf{Mean}&\textbf{Std}\\
    \hline\addlinespace[2pt]
    \multirow{4}{*}{[30\degree, 90\degree] \& [210\degree, 270\degree]}
    &SSIM& 0.9904 &0.005973\\
    &PSNR& 38.18 dB&1.280 dB\\
    &SSIM (sinogram)& 0.9748 & 0.003419\\
    &PSNR (sinogram)& 38.30 dB & 0.4866 dB\\
    \hline
    \multirow{4}{*}{[60\degree, 120\degree] \& [240\degree, 300\degree]}
    &SSIM& 0.9907&0.007775\\
    &PSNR& 38.24 dB&1.443 dB\\
    &SSIM (sinogram)& 0.9756&0.001428\\
    &PSNR (sinogram)& 38.58 dB&0.4594 dB\\
    \hline
    \multirow{4}{*}{\parbox{4.8cm}{[60\degree, 90\degree] \& [130\degree, 160\degree] \& [240\degree, 260\degree] \& [320\degree, 340\degree]}}
    &SSIM& 0.9925&0.00573\\
    &PSNR& 38.59 dB&1.829 dB\\
    &SSIM (sinogram)& 0.9772&0.001166\\
    &PSNR (sinogram)& 39.34 dB&0.3157 dB\\
    \hline \hline
    \end{tabular}
\end{table}

\begin{figure*}[ht]
    \centering
    \includegraphics[width=\textwidth]{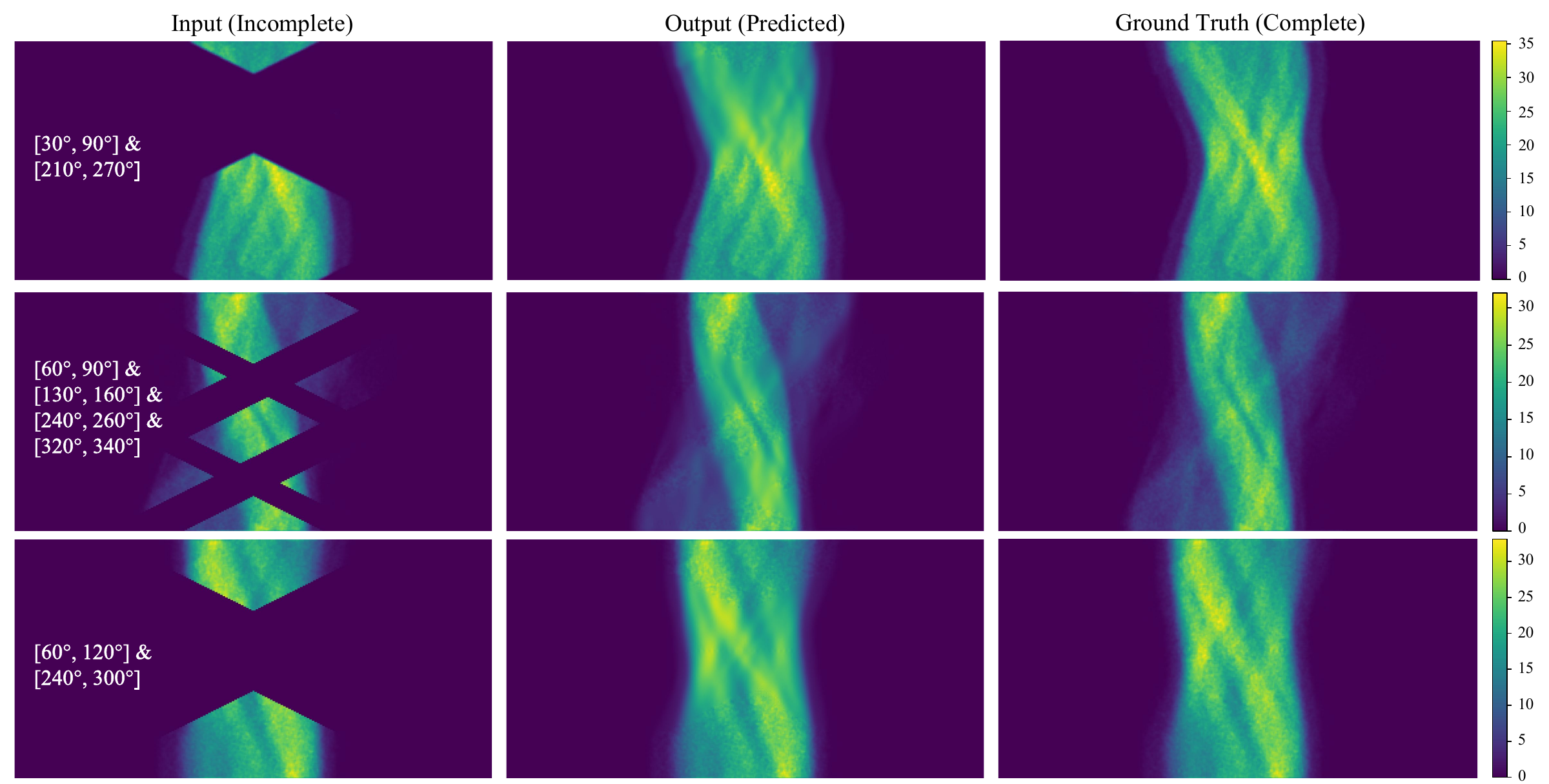}
    \vspace{-1cm}
    \caption{Comparison of PET sinogram reconstruction results after training (metrics are shown in Table~\ref{tab:results}). Each row displays three images: the left shows the incomplete input sinogram with missing angular data, the middle shows the model-predicted complete sinogram, and the right shows the complete real sinogram. And each row corresponds to a different angular loss range, as indicated in the first column. }
    \label{fig:pet_reconstruction_results}
\end{figure*}

Figure~\ref{fig:pet_reconstruction_results} demonstrates the model's capability in reconstructing incomplete sinograms. After 40 rounds of training, the model can effectively recover complete sinogram structures from inputs with missing angular data. From the figure, it can be observed that although the input sinogram (left) has large-scale data loss, the model's predicted sinogram (middle) successfully restores a structure and signal distribution highly similar to the real sinogram (right). This indicates that our proposed two-stage restoration framework can effectively learn the potential structures and features in sinograms, enabling accurate reconstruction even in cases of severe data loss.

\begin{table}[htbp]
    \centering
    \caption{Ablation studies on the contribution of each component in this framework. The results are based on the test set in the geometry space with the lost angle range of [60\degree, 90\degree] \& [130\degree, 160\degree] \& [240\degree, 260\degree] \& [320\degree, 340\degree].}
    \label{tab:ablation}
    \begin{tabular}{l l l}
    \hline \hline \addlinespace[2pt]
    \textbf{Model} & \textbf{SSIM} & \textbf{PSNR} \\
    \hline\addlinespace[2pt]
    OSEM only & 0.5608 & 12.90\,dB \\
    OSEM + 3D U-Net & 0.9293  & 31.74\,dB\\
    Attention U-Net + OSEM & 0.9860  & 36.29\,dB\\
    Attention U-Net (single channel) + OSEM & 0.9816 & 34.22\,dB \\
    UNETR + OSEM & 0.9764 & 35.61\,dB\\
    TransUNet + OSEM & 0.9770 & 35.85\,dB\\
    Attention U-Net + OSEM + 3D U-Net & 0.9918 & 38.48\,dB \\
    Attention U-Net + OSEM + 3D U-Net + Diffusion & 0.9925 & 38.59\,dB\\
    \hline \hline
    \end{tabular}
\end{table}

Table~\ref{tab:ablation} also provides a quantitative contribution of each component in this framework. The first stage U-Net is the most critical that contributes the most to the final performance. If only OSEM and the second stage are used, the SSIM drops from 0.9925 to 0.9293. The second stage in the geometry space also provides a significant boost, improving the SSIM from 0.9860 to 0.9925 when added on top of the first-stage architectures. Attention U-Net achieves the best performance 
with an SSIM of 0.9860 and a PSNR of 36.29\,dB, outperforming UNETR 
(0.9764 / 35.61\,dB) and TransUNet (0.9770 / 35.85\,dB). The advantage 
is consistent but modest, which we attribute to the relatively limited 
dataset size that may disadvantage the larger transformer-based models.

\begin{figure}[ht]
    \centering
    \includegraphics[width=\textwidth]{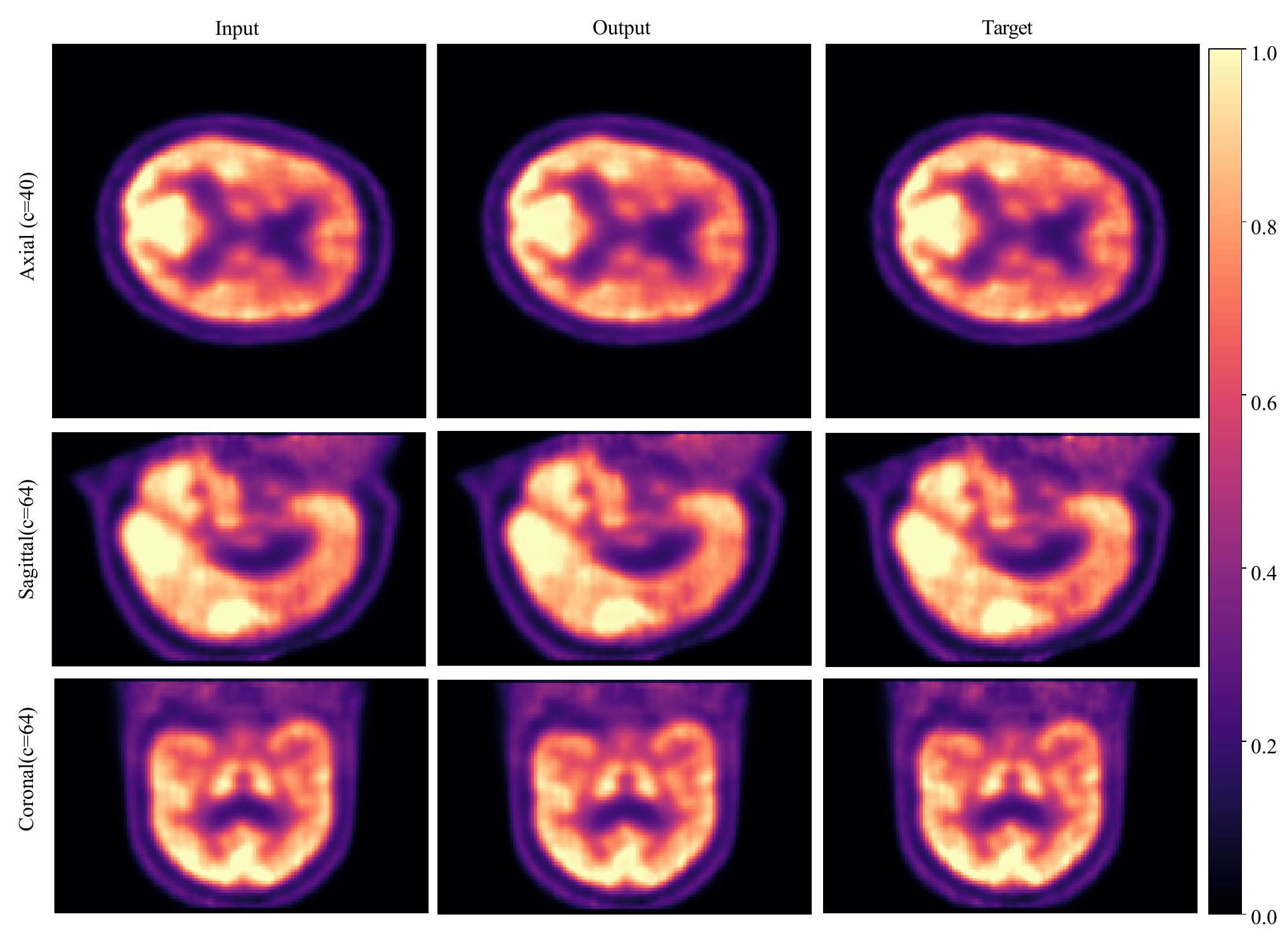}
    \vspace{-1cm}
    \caption{Comparison of original PET brain image (right) with PET image refined from the raw images reconstructed from predicted sinogram (center) with the SSIM of 0.9946 and PSNR of 39.18 dB. The left is the raw image reconstructed from the predicted sinogram. The lost angle range is [60\degree, 90\degree] \& [130\degree, 160\degree] \& [240\degree, 260\degree] \& [320\degree, 340\degree] and the sample is from Alzheimer's disease subject. 
    }
    \label{fig:compare_reconstruction_restoration}
\end{figure}

Figure~\ref{fig:compare_reconstruction_restoration} shows the PET image quality generated from the reconstructed sinogram. 
It can be seen that the reconstructed image successfully preserves key anatomical structures and tracer distribution features from the original image. Particularly in the cerebral cortex and basal ganglia regions, the reconstructed image clearly preserves the boundaries and contrast of high uptake areas. 


\subsection*{Comparison with existing methods}

Liu et al.~\cite{liu2019partial} was the first work on recovering incomplete-ring PET, where they used a supervised residual U-Net in the image domain, achieving SSIM
values of 0.993--0.998, but their experiments were limited to mild data
loss of 3/5/7 missing blocks out of 56 blocks in total and on only 20 simulated brain volumes. We used our simulation pipeline to estimate that this corresponds to a data loss of approximately 10.4--24.5\%. Shan et
al.~\cite{shan2024dip} proposed an unsupervised deep image prior (DIP)
method, with the PSNR of 26--27\,dB under a comparable missing level. However, this method should be re-optimised for every individual scan, which limits its clinical throughput.
Makkar et al.~\cite{makkar2024inr} extended this direction using
implicit neural representations (INR). They also validated the methods on three different situations, including digital brain phantoms, MC-simulated Derenzo phantom, and experimentally acquired hot-rod phantom data. 
In contrast, our two-stage framework is validated under about 50\% coincidence loss, which is double the severity addressed by prior CNN-based methods. We also have PSNR of 38.30--38.59\,dB and SSIM of 0.9907--0.9925 on a dataset of 613 brain volumes with healthy, AD, and MCI samples, much larger than the 20 phantoms used in previous works ~\cite{liu2019partial}. Most previous methods 
have also relied on simulated data~\cite{liu2019partial, shan2024dip}, 
though Makkar et al.~\cite{makkar2024inr} did validate on experimentally 
acquired phantom data.

\begin{table*}[htbp]
\centering
\caption{Comparison of deep learning methods for incomplete-ring PET reconstruction.
All PSNR and SSIM values are in the image domain.
MC = Monte Carlo simulation}
\label{tab:comparison}
\renewcommand{\arraystretch}{1.5}
\begin{tabular}{llllcc}
\hline
\textbf{Reference} & \textbf{Method} & \textbf{Subjects}  & \textbf{Missing level} & \textbf{PSNR} & \textbf{SSIM} \\
\hline

Liu el al.~\cite{liu2019partial}
  & Residual U-Net
  & 20 phantoms
  & \makecell[l]{10.4\%-24.5\%}
  & ---
  & \makecell[c]{0.993-0.998} \\

Shan et al.~\cite{shan2024dip}
  & \makecell[l]{DIP + contourlet \\ (unsupervised)}
  & \makecell[l]{Phantoms \\ + 1 rat}
  & \makecell[l]{24.5\%-35.9\%}
  & \makecell[c]{27.15 / 26.19}
  & \makecell[c]{0.939 / 0.931} \\

Makkar et al.~\cite{makkar2024inr}
  & \makecell[l]{INR \\ (unsupervised)}
  & \makecell[l]{20 phantoms \\ + 2 phantoms}
  & \makecell[l]{---}
  & \makecell[c]{29.91 / 31.73 \\ / 35.49$^{a}$}
  & \makecell[c]{0.94 / 0.95 \\ / 0.98$^{a}$} \\

\textbf{Fang \& Zhou }
  & \makecell[l]{\textbf{U-Net +} \\ \textbf{Diffusion}}
  & \makecell[l]{\textbf{613 volumes}}
  & \makecell[l]{\textbf{48.1\%-52.3\%}}
  & \textbf{38.30-38.59}
  & \textbf{0.9907-0.9925} \\

\hline
\multicolumn{6}{l}{$^{a}$ Brain phantom, Derenzo phantom (with prior),  hot-rod phantom, respectively.} \\
\end{tabular}
\end{table*}

\section{Conclusion and Discussion}
\label{chap:conclusion}

From the experimental results, we demonstrate that our two-stage reconstruction can effectively recover PET images from incomplete ring geometries. Compared to previous methods validated on at most 20 phantoms~\cite{liu2019partial, makkar2024inr}, our framework is trained and evaluated on 613 brain volumes covering healthy, 
AD, and MCI subjects, providing a substantially more robust 
statistical basis for the reported metrics.
First, our model performs remarkably well despite operating with approximately 50\% missing data and $1/3$ missing detectors, in which case direct reconstruction without restoration will cause a chaos. 
But our approach was still able to achieve PSNR values above 38 dB and SSIM scores above 0.99 for images, indicating successful ability to recover missing information of key structural features.

While the total amount of missing data remains constant, the specific distribution of the missing affects the quality of the recovery. The model extrapolates the missing parts with the help of existing data, and when the sinogram contains an entire missing region, the model will infer the missing parts from farther away, which will lead to a decrease in the quality of recovery, as shown in the second and third rows of Figure~\ref{fig:pet_reconstruction_results}. However, when the missing and intact parts are interleaved, the quality of recovery improves because the model is able to extrapolate from a closer and thus more accurate location.

Additionally, we also found that attention mechanism is essential for the model to learn the relationship between different parts of the sinogram regardless of distances between them.
Comparing the second and third rows of Figure~\ref{fig:pet_reconstruction_results}, we can observe different patterns of missing data. In the third row, the missing portions are primarily concentrated in the middle region of the sinogram, with relatively balanced data preserved in both the upper and lower sections. In contrast, the second row's missing data extends closer to the upper part of the sinogram, with only a narrow band of preserved data in the upper region.  
So if the model can only capture short distance relationships, the second row will be more difficult to recover than the third row because the model has to extrapolate the missing parts from farther away.
However, the results from Table~\ref{tab:results} show that the PSNR and SSIM of the second row are actually quite similar to those of the third row. This indicates that the attention mechanism can help the model learn the relationship between different parts of the sinogram regardless of distances between them, which is very important for the model to leverage the existing data effectively to recover the missing parts.
Six-fold cross-validation confirmed PSNR of 38.59\,dB and SSIM 
of 0.9925 on the test set.

Despite encouraging results, our study has several important limitations. First, the Monte-Carlo data were generated under idealised detector conditions that do not yet reflect the full range of crystal-to-crystal variations observed in commercial scanners, which may curb the method’s immediate transferability. 
Second, validation was performed on brain PET scans covering 
healthy controls, AD, and MCI subjects; performance on other 
organs or on more aggressive pathological uptake patterns 
(e.g., tumours with focal high uptake) remains unknown. 
The diffusion module could in principle suppress subtle lesions 
or introduce artificial features on such out-of-distribution 
inputs, and validation on a broader range of pathologies is 
therefore a priority for future work.
Third, our network processes 2-D slices augmented by neighbouring information, a compromise that can sacrifice some genuinely three-dimensional context. 
Fourth, although SimSET has been shown to be consistent with GATE to 
within 4--11\%~\cite{poon2015simset}, it has not yet achieved the 
same community acceptance. So migrating to GATE remains a priority 
for future work.
Finally, the evaluation covered only one scanner radius (253.71 mm); further work is required to understand the model’s behaviour across different geometries and partial-ring layouts.

Looking toward future improvements, several avenues can close these gaps. 
The most important work is extending our validation to a broader range of pathological uptake patterns. Focal lesions and tumours should be prioritized as in-beam therapy monitoring is one of the most common applications of incomplete-ring PET. This situation usually involves patients 
with active tumours whose uptake patterns differ significantly 
from the healthy and neurodegenerative cases studied here.
Liu et al.~\cite{liu2019partial} additionally evaluated 
recovery coefficients for grey matter, white matter, and 
two artificially inserted lesions. We plan to use similar data in future work to help evaluate the hallucination risk of our method and to better understand its performance on pathological cases.
Collecting list-mode data on physical PET systems—ideally including scanners with deliberately inactive modules—will provide the experimental evidence needed for clinical translation. 
To preserve inter-slice relationships without incurring a full 3-D memory cost, hybrid approaches that add recurrent or transformer-style attention across slices merit exploration. 
Introducing physics-informed constraints during training could further narrow the gap between simulated and real acquisitions. 
Migrating the workflow to a GATE-based simulation pipeline would align the study with established practice and enhance credibility.
Finally, larger and more diverse datasets may unlock the full potential of transformer-driven architectures such as UNETR or TransUNet, which showed only marginal advantages in the present, more limited setting.

\section{Declarations}
\subsection*{Conflict of Interest}
The authors declare that they have no conflict of interest.

\subsection*{Data Availability}
Data used in the preparation of this article were obtained 
from the Alzheimer's Disease Neuroimaging Initiative (ADNI) 
database (adni.loni.usc.edu). Researchers may apply for 
access to ADNI data at \url{https://adni.loni.usc.edu/data-samples/access-data/}.

\subsection*{Ethics statement}
Brain PET images used as phantom references in this study 
were obtained from the Alzheimer's Disease Neuroimaging 
Initiative (ADNI) database (adni.loni.usc.edu). The ADNI 
study was conducted in accordance with the Good Clinical 
Practice guidelines and applicable federal regulations. 
Written informed consent was obtained from all ADNI 
participants or their authorised representatives at each 
participating site.

\subsection*{Competing interests}
The authors declare that they have no competing interests.

\subsection*{Consent for publication}
Not applicable.

\section{Acknowledgments}
I would like to express my sincere gratitude to senior students, Xiaoyu Chen for insightful discussions and suggestions.
The conclusions and analyses presented in this publication were produced using the following software: Python (Guido van Rossum 1986) \cite{10.5555/1593511}, Opencv (Intel, Willow Garage, Itseez) \cite{itseez2015opencv}, Scipy (Jones et al. 2001) \cite{2020SciPy-NMeth}, PyTorch  (Meta AI September 2016) \cite{NEURIPS2019_9015}, Matplotlib (Hunter 2007) \cite{Hunter:2007} and Seaborn \cite{Waskom2021}.
Data collection and sharing for this project was funded by the Alzheimer's Disease Neuroimaging Initiative
(ADNI) (National Institutes of Health Grant U01 AG024904) and DOD ADNI (Department of Defense award
number W81XWH-12-2-0012). ADNI is funded by the National Institute on Aging, the National Institute of
Biomedical Imaging and Bioengineering, and through generous contributions from the following: AbbVie,
Alzheimer’s Association; Alzheimer’s Drug Discovery Foundation; Araclon Biotech; BioClinica, Inc.; Biogen;
Bristol-Myers Squibb Company; CereSpir, Inc.; Cogstate; Eisai Inc.; Elan Pharmaceuticals, Inc.; Eli Lilly and
Company; EuroImmun; F. Hoffmann-La Roche Ltd and its affiliated company Genentech, Inc.; Fujirebio; GE
Healthcare; IXICO Ltd.; Janssen Alzheimer Immunotherapy Research \& Development, LLC.; Johnson \&
Johnson Pharmaceutical Research \& Development LLC.; Lumosity; Lundbeck; Merck \& Co., Inc.; Meso
Scale Diagnostics, LLC.; NeuroRx Research; Neurotrack Technologies; Novartis Pharmaceuticals
Corporation; Pfizer Inc.; Piramal Imaging; Servier; Takeda Pharmaceutical Company; and Transition
Therapeutics. The Canadian Institutes of Health Research is providing funds to support ADNI clinical sites
in Canada. Private sector contributions are facilitated by the Foundation for the National Institutes of Health
(www.fnih.org). The grantee organization is the Northern California Institute for Research and Education,
and the study is coordinated by the Alzheimer’s Therapeutic Research Institute at the University of Southern
California. ADNI data are disseminated by the Laboratory for Neuro Imaging at the University of Southern
California.

\appendix

\section*{Appendix}
\label{sec:appendix}

The parameters of the five-channel U-Net network structure are shown in Table~\ref{tab:unet_architecture}. We also have skip connections between the encoder and decoder blocks, which are not shown in the table. 

\begin{table}[htbp]
    \centering
    \caption{Five-channel Attention U-Net}
    \vspace{0.1cm}
    \label{tab:unet_architecture}
    \begin{tabular}{@{}l c c@{}}
        \toprule
        \textbf{Stage / block(s)} & \textbf{Main ops (per level)} & \textbf{$C_\text{out}$} \\
        \midrule
        Input & --- & 5 \\[3pt]
        
        \multirow{2}{*}{Encoder $\times$4} 
          & [Conv $3\times3$ + BN + ReLU]$\times2$ & \multirow{2}{*}{64, 128, 256, 512}\\
          & MaxPool $2\times2$                        & \\[3pt]
        
        Bottleneck & [Conv $3\times3$ + BN + ReLU]$\times2$ & 1024 \\[3pt]
        
        \multirow{2}{*}{Decoder $\times$4}
          & Upsample ($\uparrow$2) + skip concat      & \multirow{2}{*}{512, 256, 128, 64}\\
          & [Conv $3\times3$ + BN + ReLU]$\times2$  & \\[3pt]
        
        Output & Conv $1\times1$ & 5 \\
        \bottomrule
        \end{tabular}
        
  \end{table}

All hyperparameters required to replicate the run are collected in Table~\ref{tab:num-hparams}.

\begin{table*}[htbp]
  \centering
  \label{tab:num-hparams}
  
  \caption{Numerical hyper-parameters for the stage-2 run}
  \vspace{0.1cm}
  \begin{minipage}[t]{0.4\linewidth}\centering
    \begin{tabular}{@{}l r@{}}
      \toprule
      \textbf{Hyper-parameter} & \textbf{Value}\\
      \midrule
      \multicolumn{2}{@{}l}{\emph{Data}}\\
      \quad batch\_size & 9\\
      \quad l\_resolution (=r) & 64\\[2pt]
      \multicolumn{2}{@{}l}{\emph{U-Net}}\\
      \quad in\_channel & 1\\
      \quad out\_channel & 1\\
      \quad inner\_channel & 64\\
      \quad res\_blocks & 3\\[2pt]
      \multicolumn{2}{@{}l}{\emph{diffusion model}}\\
      \quad in\_channel & 2\\
      \quad out\_channel & 1\\
      \quad inner\_channel & 32\\
      \quad res\_blocks & 3\\
      \bottomrule
    \end{tabular}
  \end{minipage}%
  \begin{minipage}[t]{0.4\linewidth}\centering
    \begin{tabular}{@{}l r@{}}
      \toprule
      \textbf{Hyper-parameter} & \textbf{Value}\\
      \midrule
      \multicolumn{2}{@{}l}{\emph{Channel multipliers}}\\
      \quad levels & 1, 2, 3, 4\\[2pt]
      \multicolumn{2}{@{}l}{\emph{Diffusion process}}\\
      \quad image\_size & 128\\
      \quad $T_{\text{train}}$ & 2 000\\
      \quad $T_{\text{val}}$ & 10\\
      \quad $\beta_{\mathrm{start}}$ & $1\times10^{-6}$\\
      \quad $\beta_{\mathrm{end,train}}$ & 0.01\\
      \quad $\beta_{\mathrm{end,val}}$ & 0.5\\[2pt]
      \multicolumn{2}{@{}l}{\emph{Optimisation}}\\
      \quad n\_iter & 600 000\\
      \quad learning rate & 0.0002\\
      \quad EMA decay & 0.9999\\
      \bottomrule
    \end{tabular}
  \end{minipage}
\end{table*}


\bibliographystyle{unsrt}
\bibliography{sinogram}

\end{document}